\documentclass[letterpaper, 10pt, conference]{ieeeconf}  

\IEEEoverridecommandlockouts                              
                                                          
\usepackage[utf8]{inputenc}
\usepackage[english]{babel}
\usepackage{amsmath}
\usepackage{amssymb}
\usepackage{graphicx}
\usepackage{subcaption}
\usepackage{hyperref}
\usepackage{url}
\usepackage{xcolor}

\usepackage{dcolumn,booktabs}
\newcolumntype{d}[1]{D{.}{.}{#1}} 

\title{\LARGE \bf Inertial-Only Optimization for Visual-Inertial Initialization}
\author{Carlos Campos, José M.M. Montiel and Juan D. Tardós
\thanks{This work was supported in part by the Spanish government under grants  PGC2018-096367-B-I00 and DPI2017-91104-EXP, the Aragón government under grant DGA\_T45-17R, and by Huawei under grant HF2017040003.}
\thanks{The authors are with Instituto de Investigación en Ingeniería de Aragón (I3A), Universidad de Zaragoza, Spain   {\tt\small campos@unizar.es; josemari@unizar.es; tardos@unizar.es}}%
}

\begin{document}

\thispagestyle{empty}
\newpage
\onecolumn
\begin{center}
This paper has been accepted for publication in 2020 International Conference on Robotics and Automation (ICRA).
\vspace{0.75cm}\\
DOI: \\ 
IEEE Xplore: \\
\vspace{1.25cm}
\end{center}
©2020 IEEE. Personal use of this material is permitted. Permission from IEEE must be obtained for all other uses, in any current or future media, including reprinting/republishing this material for advertising or promotional purposes, creating new collective works, for resale or redistribution to servers or lists, or reuse of any copyrighted component of this work in other works.
\twocolumn

\maketitle
\thispagestyle{empty}
\pagestyle{empty}

\begin{abstract}

We formulate for the first time visual-inertial initialization as an optimal estimation problem, in the sense of maximum-a-posteriori (MAP) estimation. This allows us to properly take into account IMU measurement uncertainty, which was neglected in previous methods that either solved sets of algebraic equations, or minimized ad-hoc cost functions using least squares. Our exhaustive initialization tests on EuRoC dataset show that our proposal largely outperforms the best methods in the literature, being able to initialize in less than 4 seconds in almost any point of the trajectory, with a scale error of 5.3\% on average. This initialization has been integrated into ORB-SLAM Visual-Inertial boosting its robustness and efficiency while maintaining its excellent accuracy. 
\end{abstract}

\section{Introduction}
Simultaneous Localization and Mapping (SLAM) techniques allow robots and AR/VR systems to be aware of their environments, while locating themselves in the reconstructed scene. Visual-inertial SLAM with a single monocular camera and a low-cost Inertial Measurement Unit (IMU) sensor, offers a small, compact and low power solution for most applications. IMU sensors measure acceleration and angular velocity, providing robustness against fast motion or challenging environments, and allowing to retrieve the true scale of the environment, which would remain unknown in a pure monocular system.

However, to start using them, some parameters need to be estimated in an initialization process. These are scale, gravity direction, initial velocity, and accelerometer and gyroscope biases. A wrong initialization would lead to poor convergence, as well as inaccurate estimation of all other variables. In addition, a fast initialization is as important as an accurate one, because as long as IMU is not initialized, visual-inertial SLAM cannot be performed.

Previous works on visual-inertial initialization can be classified in joint and disjoint (or loosely coupled) estimation methods. Joint visual-inertial initialization was pioneered by Martinelli \cite{martinelli2014closed}, who proposed a closed-form solution to jointly retrieve scale, gravity, accelerometer bias and initial velocity, as well as visual features depth. This method was built on the assumption that camera poses can be roughly estimated from IMU readings. The method tracks several points in all the images, and builds a system of equations stating that the 3D point coordinates, as seen from any camera pair, should be the same, that is solved by linear least squares. This work was extended by Kaiser et al. \cite{kaiser2017simultaneous} building a similar linear algebraic system that is solved using non-linear least squares, to also find gyroscope bias and to take gravity magnitude into account. The capability to find accurate initial solutions in 2 seconds was shown in simulations.

Crucially, the original and modified methods ignore IMU noise properties, and minimize the 3D error of points in space, and not their reprojection errors, that is the gold-standard in feature-based computer vision. Our previous work \cite{campos2019fast} shows that this results in large unpredictable errors, that can be corrected by adding two rounds of Visual-Inertial Bundle Adjustment (VI-BA), together with two tests to detect and discard bad initializations. This renders the method usable, obtaining on the public EuRoC dataset \cite{burri2016euroc} joint visual-inertial initializations in 2 seconds with scale error around 5\%. However, the method only works in 20\% of the trajectory points. Such a low initialization recall can be a problem for AR/VR or drone applications where the system is desired to be launched immediately.

Disjoint visual-inertial initialization is based on the solid assumption that the up-to-scale camera trajectory can be estimated very accurately from pure monocular vision, and then use this trajectory to estimate the inertial parameters. As modern visual-odometry and visual SLAM systems perform local bundle adjustment and provide trajectories with much higher precision than  IMU integration, this trajectory uncertainty can be safely ignored while estimating the inertial parameters. This idea was pioneered by Mur-Artal and Tardós in ORBSLAM-VI \cite{mur2017visual}, and later adopted by Qin et al. in VINS-Mono \cite{qin2017robust} \cite{qin2018vins}. In both cases, inertial parameters are found in different steps by solving a set of linear equations using least-squares. In \cite{mur2017visual} a linear system is built by eliminating the velocities for each frame. However, after these algebraic manipulations, the errors to be minimized are meaningless and unrelated to sensor noise properties. In order to obtain accurate estimations, including accelerometer bias, the method requires 15 seconds for initialization. In \cite{qin2018vins} accelerometer bias is assumed to be zero, requiring only 1-2 seconds to initialize, depending on the motion. In both methods, IMU measurements are manipulated and mixed in the same linear system, where the residuals of all equations are considered with the same weight, ignoring sensor uncertainties. In addition, the different inertial parameters are solved separately in different steps, not all at once, ignoring the correlations between them. All this leads to an estimation which is not optimal in the sense of maximum-a-posteriori (MAP) estimation.

We propose a novel disjoint visual-inertial initialization method by formulating it as an optimal estimation problem, in the sense of MAP estimation. For this, we build on the excellent work of Forster et al. \cite{forster2015imu} that allows to preintegrate IMU readings and, taking into account the probabilistic characterization of sensor noises, properly compute the covariances of the preintegrated terms. Assuming that the error of the monocular SLAM trajectory is negligible compared with the IMU errors, we derive a very efficient MAP estimator for inertial-only parameters, and use it to initialize a visual-inertial SLAM system. The main contributions of our work are:

\begin{itemize}
\item The formulation of the visual-inertial initialization as an inertial-only optimal estimation problem, in the sense of MAP estimation, taking properly into account the probabilistic model of IMU noises.
\item We solve for all inertial parameters at once, in a single step, avoiding the inconsistencies derived from decoupled estimation. This makes all estimations jointly consistent. 
\item We do not make any assumptions about initial velocity or attitude, which makes our method suitable for any initialization case. 
\item We do not assume IMU biases to be zero, instead we code the known information about them as probabilistic priors that are exploited by our MAP estimation.
\end{itemize}

In the next section we present the theory and in-depth details behind our proposal. Later, we evaluate and compare it against the best examples of joint and disjoint initialization methods, proving to outperform them.

\section{Maximum-A-Posteriori Initialization}

The gold-standard method for feature-based visual-inertial SLAM is visual-inertial bundle adjustment (VI-BA), that takes properly into account the noise properties in all the sensors, and obtains a maximum-a-posteriori joint estimation of all variables (see   \cite{mur2017visual} for a modern formulation using  IMU preintegration on manifold from \cite{forster2015imu}). The main limitation of VI-BA is that it requires a good seed to converge quickly and avoid getting stuck in local minima, due to its strong non-linear nature.  Joint \cite{campos2019fast} and disjoint \cite{mur2017visual} initialization methods based on least-squares estimation showed that VI-BA largely improves their initial solutions.  

Our main goal is going one step further and also use  MAP estimation in the initialization, making proper use of sensor noise models. Our novel initialization method is based on the following ideas:
\begin{itemize}
    \item Despite the non-linear nature of BA, Monocular SLAM (or visual odometry) is mature and robust enough to obtain very accurate initial solutions for structure and motion, with the only caveat that their estimations are up-to-scale.
    \item The uncertainty of visual SLAM trajectory is much smaller than the IMU uncertainties and can be ignored while obtaining a first solution for the IMU variables. So, we perform inertial-only MAP estimation, taking the up-to-scale visual SLAM trajectory as constant.
    \item Inspired on the work of \cite{strasdat2010scale}, we adopt a parametrization that explicitly represents and optimizes the scale factor of the monocular SLAM solution.
    \item Differently from \cite{mur2017visual} \cite{qin2018vins}, we jointly optimize all the IMU variables in one step, taking into account the cross-covariances between the preintegrated terms for position, and linear and angular velocities \cite{forster2015imu}.
\end{itemize}

Our initialization method can be split in three steps:
\begin{enumerate}
    \item \textbf{Vision-only MAP estimation}: Initialize and run monocular ORB-SLAM \cite{mur2015orb} for a short period (typically 2 s) using BA to obtain a vision-only MAP estimation up-to-scale. At the same time, compute IMU preintegrations between keyframes and their covariances \cite{forster2015imu}.
    \item \textbf{Inertial-only MAP estimation}: Inertial-only optimization  to align the IMU trajectory and ORB-SLAM trajectory, finding the scale, keyframes' velocities, gravity direction and IMU biases. 
    \item \textbf{Visual-inertial MAP estimation}: Use the solution from the previous step as seed for a full VI-BA to obtain the joint optimal solution. 
\end{enumerate}

After the initialization, we launch ORB-SLAM Visual-Inertial \cite{mur2017visual}, that performs local VI-BA. We have observed that scale estimation accuracy can be further improved after 5-10 seconds performing a full VI-BA or, with much lower computational cost, repeating the inertial-only optimization.  

The three initialization steps are further detailed next.

\subsection{Vision-only MAP Estimation}
We initialize pure monocular SLAM, using the same procedure as in ORB-SLAM to find the initial motion. Matching of FAST points, using ORB descriptor, is performed between two initial frames. Fundamental matrix and homography models are found and scored. The one with a higher score is used to find the initial motion and triangulate the features. Once the structure and motion are initialized, we do pure monocular SLAM for 1 or 2 seconds. The only difference from ORB-SLAM is that we enforce keyframe insertion at a higher frequency (4 Hz to 10 Hz). In that way, IMU preintegration between keyframes has low uncertainty, since integration times are very short. After this period, we have an up-to-scale map composed of ten keyframes and hundreds of points, that has been optimized using BA by the ORB-SLAM mapping thread. 

The up-to-scale keyframe poses are transformed to the body (or IMU) reference using visual-inertial calibration. These body poses are denoted as $\mathbf{\bar{T}}_{0:k}=[\mathbf{R},\mathbf{\bar{p}}]_{0:k}$, where $\mathbf{R}_i \in \text{SO}(3)$ is rotation matrix from $i$-th body to world reference, and $\mathbf{\bar{p}_i}\in \mathbb{R}^3$ is the up-to-scale position of $i$-th body.

\subsection{Inertial-only MAP Estimation}
The goal of this step is to obtain an optimal estimation of the inertial parameters, in the sense of MAP estimation, using the up-to-scale trajectory obtained by vision. As we don't have a good guess of the inertial parameters, using at this point a full VI-BA would be too expensive and prone to get stuck in local minima, as shown in the experiments section. An intermediate solution would be to marginalize out the points to obtain a prior for the trajectory and its (fully dense) covariance matrix, and use it while optimizing the IMU parameters. We opt for a more efficient solution, considering the trajectory as fixed, and perform an inertial-only optimization. The inertial parameters to be found are:
\begin{equation}
    \mathcal{X}_k=\{s, \textbf{R}_{\text{w}g}, \mathbf{b}, \mathbf{\bar{v}}_{0:k} \}
\end{equation}
where $s\in \mathbb{R}^+$ is the scale factor of the vision-only solution, $\textbf{R}_{\text{w}g} \in \text{SO}(3)$ is the gravity direction, parameterized by two angles, such that gravity in world reference frame is expressed as $\textbf{g}=\textbf{R}_{\text{w}g}\textbf{g}_{\text{I}}$, with $\textbf{g}_{\text{I}}=(0,0,G)^{\text{T}}$ being $G$ the magnitude of gravity,  $\mathbf{b}=(\mathbf{b}^a,\mathbf{b}^g) \in \mathbb{R}^6$ are the accelerometer and gyroscope biases, and  $\mathbf{\bar{v}}_{0:k} \in \mathbb{R}^3$ the up-to-scale body velocities from first to last keyframe. 

We prefer to use up-to-scale velocities $\mathbf{\bar{v}}_i$, instead of true ones $\mathbf{v}_i = s \mathbf{\bar{v}}_i$, since it eases the initialization process. Biases are assumed constant for all involved keyframes since initialization period is just 1-2 seconds, and random walk would have almost no effect. It is worth noting that this formulation takes into account gravity magnitude from the beginning, as opposed to \cite{qin2018vins} and \cite{mur2017visual} that require a separate step to fix its value.

In our case, the only measurements used come from IMU, and are summarized in the IMU preintegrated terms defined in \cite{forster2015imu}. We denote by $\mathcal{I}_{i,j}$ the preintegration of inertial measurements between $i$-th and $j$-th keyframes, and by  $\mathcal{I}_{0:k}$ the set of IMU preintegrations between successive keyframes in our initialization window.

With the state and measurements defined, we can formulate a MAP estimation problem, where the posterior distribution is:

\begin{equation}
    p(\mathcal{X}_k \vert \mathcal{I}_{0:k}) \propto p( \mathcal{I}_{0:k} \vert \mathcal{X}_k) p(\mathcal{X}_k)
\end{equation}

\noindent where $p(\mathcal{I}_{0:k} \vert \mathcal{X}_k)$ is the likelihood distribution of the IMU measurements given the IMU states, and $p(\mathcal{X}_k)$ the prior for the IMU states. Considering independence of measurements, the likelihood can be factorized as:

\begin{equation}
    p(\mathcal{I}_{0:k} \vert \mathcal{X}_k) =  \prod_{i=1}^k p(\mathcal{I}_{i-1,i} \vert s, \mathbf{g}_{dir}, \mathbf{b}, \mathbf{v}_{i-1}, \mathbf{v}_{i})  
\end{equation}

To obtain the MAP estimator, we need to find the parameters which maximize the posterior distribution, that is equivalent to minimize its negative logarithm, thus:
\begin{multline}
\mathcal{X}_k^* = \arg \max_{\mathcal{X}_k} p(\mathcal{X}_k \vert \mathcal{I}_{0:k}) = \arg \min_{\mathcal{X}_k} 
\biggl( -\log (p(\mathcal{X}_k))  \\ 
\left.
- \sum_{i=1}^k \log \left( p(\mathcal{I}_{i-1,i} \vert s, \mathbf{g}_{dir}, \mathbf{b}, \mathbf{v}_{i-1}, \mathbf{v}_{i})  \right)
\right)  
\end{multline}

Assuming Gaussian error for IMU preintegration and prior distribution, the MAP problem is equivalent to:
\begin{equation} \label{eqn:minimization}
    \mathcal{X}_k^* = \arg \min_{\mathcal{X}_k} \left( \Vert \mathbf{r}_{\text{p}}\Vert_{\Sigma_p}^2 + \sum_{i=1}^{k} \Vert \mathbf{r}_{\mathcal{I}_{i-1,i}}\Vert_{\Sigma_{\mathcal{I}_{i-1,i}}}^2 \right)   
\end{equation}

\noindent where $\mathbf{r}_{\text{p}}$ and $\mathbf{r}_{\mathcal{I}_{i-1,i}}$ are the residual of the prior and IMU measurements between consecutive keyframes, while $\Sigma_p$ and $\Sigma_{\mathcal{I}_{i-1,i}}$ are their covariances. 

In this optimization, vision reprojection errors do not appear, only inertial residuals. As IMU measurements do not suffer from data association errors, the use of robust cost function, like the Huber norm, does not make sense, since it would slow down the optimization. 

Following \cite{lupton2012visual} and \cite{forster2015imu}, we define the inertial residual as:

\begin{equation}
\mathbf{r}_{\mathcal{I}_{i,j}}=[\textbf{r}_{\Delta \text{R}_{ij}}, \textbf{r}_{\Delta \text{v}_{ij}}, \textbf{r}_{\Delta \text{p}_{ij}}]
\end{equation}

\begin{equation}
\textbf{r}_{\Delta \text{R}_{ij}} = \text{Log}\left(  \Delta \textbf{R}_{ij}(\textbf{b}^g)^\text{T} \textbf{R}_i^\text{T} \textbf{R}_j\right)
\end{equation}

\begin{equation}
\textbf{r}_{\Delta \text{v}_{ij}} = \textbf{R}_i^\text{T} \left( s \bar{\textbf{v}}_j - s \bar{\textbf{v}}_i - \textbf{R}_{\text{w}g}\textbf{g}_{\text{I}}\Delta t_{ij}\right) - \Delta\textbf{v}_{ij}(\textbf{b}^g, \textbf{b}^a)
\end{equation}

\begin{multline}
\textbf{r}_{\Delta \text{p}_{ij}} = \textbf{R}_i^\text{T} \left( s \bar{\textbf{p}}_j - s \bar{\textbf{p}}_i - s \bar{\textbf{v}}_i \Delta t_{ij} - \frac{1}{2}\textbf{R}_{\text{w}g}\textbf{g}_{\text{I}}\Delta t_{ij}^2\right) \\ - \Delta\textbf{p}_{ij}(\textbf{b}^g, \textbf{b}^a)
\end{multline}

\noindent where $\Delta \textbf{R}_{ij}(\textbf{b}^g)$, $\Delta\textbf{v}_{ij}(\textbf{b}^g, \textbf{b}^a)$ and $\Delta\textbf{p}_{ij}(\textbf{b}^g, \textbf{b}^a)$ are preintegrated IMU measurements from $i$-th to $j$-th keyframe, which only depend on biases. These terms can be linearly updated as explained in \cite{forster2015imu}, avoiding reintegrating at each iteration. $\Delta t_{ij}$ is the time between both keyframes. $\text{Log}$ stands for the logarithm map from Lie group $\text{SO(3)}$ to its algebra $\mathfrak{so}(3)$, isomorphic to $\mathbb{R}^3$. Since we assume that biases can be considered constant during the initialization window, IMU residuals do not include random walk for biases. We assume that the residuals follow Gaussian distributions, and their covariances can be computed as proposed in \cite{forster2015imu}.

As we are optimizing in a manifold we need to define a retraction \cite{forster2015imu} to update the gravity direction estimation during the optimization:
\begin{equation}
\textbf{R}_{ \text{wg}}^{\text{new}} = \textbf{R}_{ \text{wg}}^{\text{old}} \text{Exp}(\delta \alpha_{\textbf{g}}, \delta \beta_{\textbf{g}}, 0)
\end{equation}

\noindent being $\text{Exp}(.)$ the exponential map from $\mathfrak{so}(3)$ to $\text{SO}(3)$. To guarantee that scale factor remains positive during optimization we define its update as:
\begin{equation}
s^{\text{new}} = s^{\text{old}} \exp{(\delta s)}
\end{equation}

Biases and velocities are updated additively. If we define $\delta \mathbf{g}_{dir}=(\delta \alpha_{\mathbf{g}}, \delta \beta_{\mathbf{g}})$, the inertial parameters updates used during optimization are $(\delta s, \delta \mathbf{g}_{dir}, \delta \mathbf{b}^g, \delta \mathbf{b}^a, \{\delta \mathbf{\bar{v}}_i\})$. Derivatives of IMU residuals w.r.t. these parameters can be found in the appendix.

\begin{figure}
  \centering
  \includegraphics[width=0.48\textwidth]{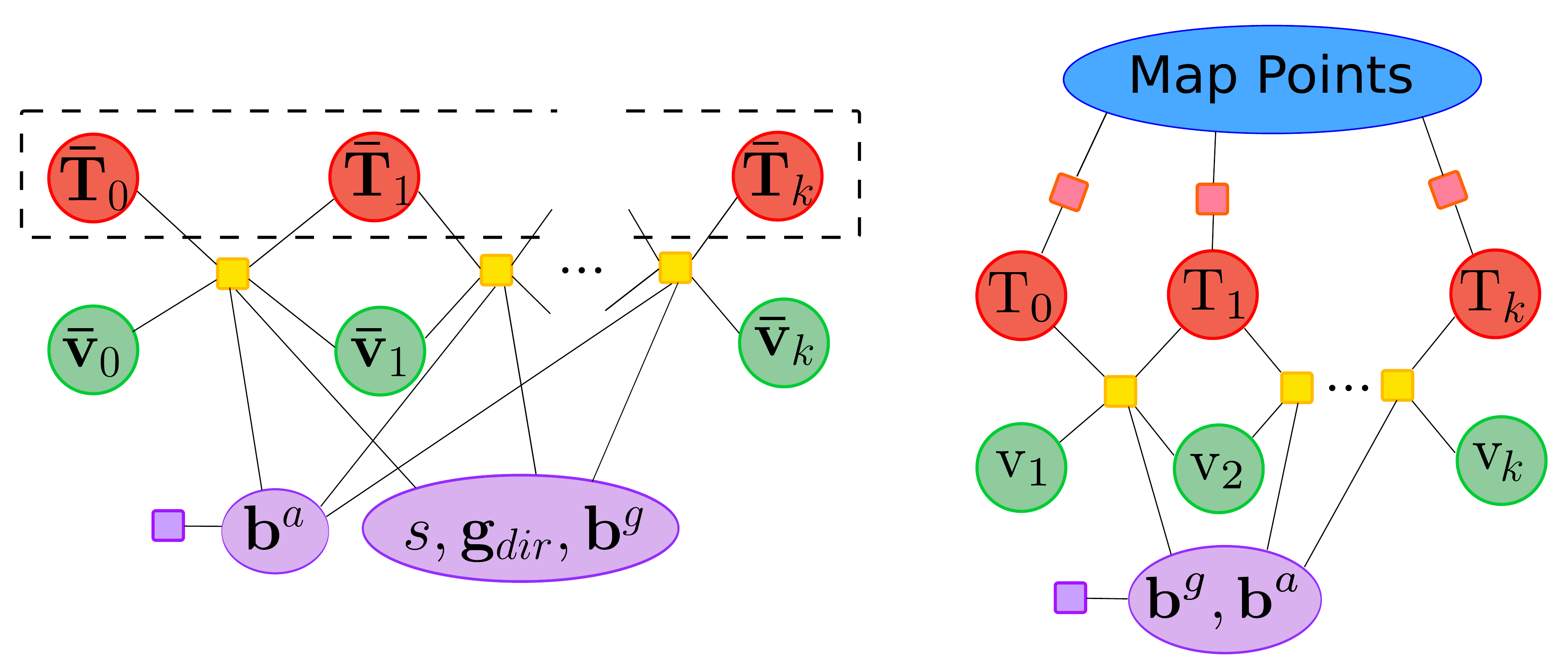}
  \caption{Underlying graph representation of the inertial-only optimization (\textbf{left}) and the first visual-inertial Bundle Adjustment (\textbf{right}). Yellow boxes represent IMU residuals, red boxes stand for reprojection error, while the purple one represents prior information for accelerometer bias. Dashed lines point out fixed variables (keyframes poses for inertial-only optimization)}
  \label{fig:OnlyInertialOptimization}
\end{figure}

The final optimization problem, represented in figure \ref{fig:OnlyInertialOptimization}, is implemented and solved using g2o C++ library \cite{kummerle2011g}, using analytic derivatives and  Levenberg-Marquardt algorithm. 

As it is well known in the literature, gravity and accelerometer bias tends to be coupled, being difficult to distinguish in most cases. To avoid that problem, some techniques neglect accelerometer bias during the initialization assuming a zero value \cite{qin2018vins}, while others wait for a long time to guarantee that it is observable \cite{mur2017visual}. Here we adopt a sound and pragmatic approach: we include $\textbf{b}^a$ as a parameter to be optimized, but adding a prior residual for it: $\mathbf{r}_{p}=\Vert \mathbf{b}^a \Vert^2_{\Sigma_p}$. If the motion performed does not contain enough information to estimate the bias, the prior will keep its estimation close to zero. If the motion makes $\textbf{b}^a$ observable, its estimation will converge towards its true value. A prior for  $\textbf{b}^g$ is not needed as it is always well observable from keyframe orientations and gyroscope readings. 

Since we have to solve a non-linear optimization problem, we need an initial guess for inertial parameters. Hence, we initialize biases equal to zero, while gravity direction is initialized along the average of accelerometer measurements, as accelerations are usually much smaller than gravity. 
 


The scale factor needs to be initialized sufficiently close to its true value to guarantee  convergence, but we do not have any initial guess. Taking advantage of our very efficient inertial-only optimization (5ms), we launch the optimization with three initial scale values, that correspond to median scene depth of 1, 4 and 16 meters, keeping the solution that provides the lowest residual as defined in equation \ref{eqn:minimization}. Our results show that, using this range of scale values, our method is able to converge in a wide variety of scenes. 

At the end of the optimization, the frame poses and velocities and the 3D map points are scaled with the scale value found, and are rotated to align the $z$ axis with the estimated gravity direction. IMU preintegration is repeated with the new bias estimations, aiming to reduce future linearization errors.

\subsection{Visual-Inertial MAP Estimation}


Inertial-only optimization provides an estimation accurate enough to be used as seed for a first joint visual-inertial Bundle Adjustment, ensuring its convergence. In this optimization, shown also in figure \ref{fig:OnlyInertialOptimization}, pure inertial parameters like $\mathbf{g}_{dir}$ and $s$ do not appear, but they are implicitly included in keyframe poses. Compared with \cite{campos2019fast}, this step replaces the BA1\&2 steps. In fact, the optimization is exactly the same, it only differs in the initial seed, which previously was computed solving a linear system, and now is computed by means of a MAP estimator. A similar optimization is also done in VINS-Mono initialization, before launching VI odometry. 

In the literature, there are several  proposed tests to determine if an initialization is successful or not. In \cite{campos2019fast}, observability of the optimization problem and consensus between different sets of measurements are checked. In contrast, VINS-Mono checks that estimated gravity magnitude has an error lower than 10\%, and IMU readings have enough variance. Here, we propose to discard initializations whose mean acceleration is below some threshold (0.5\% of gravity). This discards only the worst attempts, with almost constant velocity, which are not observable \cite{martinelli2014closed}.

We remark that all initialization steps are performed in a parallel thread, without having any effect on the real time tracking thread. Once the optimization is finished, the system is already initialized, and we switch from visual to visual-inertial SLAM.

\section{Experimental Results}

To analyze the capability to initialize under different sensor trajectories, we run an exhaustive initialization test. We launch an initialization every 0.5 seconds (one out of 10 frames) in every trajectory of the EuRoC dataset, what results on testing 2248 different initialization trajectories. To compare, we run the same exhaustive test with the joint initialization method of \cite{campos2019fast} and the loosely coupled initialization of VINS-Mono \cite{qin2018vins}, using the software provided by the authors. As a baseline, we also try to initialize  using only visual-inertial bundle adjustment with the same initial guesses for gravity direction, velocities and biases, and the same three initial values for scale, keeping the solution with smaller residual. 

The performance is measured in terms of the scale error before and after applying full VI-BA.
To measure scale factor, and thus scale error, we align the initialization and ground-truth trajectories using Horn alignment, such that for an instant $t$, the estimated $\mathbf{\hat{p}}(t)$ and ground-truth $\mathbf{p}_{\text{GT}}(t)$ trajectories are related by:

\begin{equation}
    \mathbf{\hat{p}}(t) = \mathbf{T} \oplus \mathbf{p}_{\text{GT}}(t) \quad \text{where} \quad \mathbf{T} \in \text{Sim}(3)
\end{equation}

We also report the duration of the initialization trajectory, denoted as $t_{Init}$, as well as the total time, $t_{Tot}$, until a successful initialization is achieved. For all  methods, if a bad initialization is detected, a new one is attempted with the next batch of available data. This, together with the time needed for visual initialization, makes $t_{Tot} \geq t_{Init}$.

\begingroup
\begin{table*}[t]
\scriptsize
\centering
\caption {\label{tab:ResultsInit} Results of exhaustive initialization attempts every 0.5s in EuRoC dataset. The first two blocks compare our proposal with the best joint initialization method \cite{campos2019fast} using trajectories of $\sim 2$ seconds ($t_{Init}$), while the last two blocks compare it with the loosely-coupled initialization of VINS-mono \cite{qin2018vins} using trajectories of $\sim 1.3$ seconds.}
\setlength\tabcolsep{2.0pt}

\begin{tabular}{|c||cc|cc|cc|cc|c||cc|cc|cc|cc|c|}
\hline
  
 {} & \multicolumn{4}{c|}{Joint Initialization  \cite{campos2019fast} }  & \multicolumn{4}{c|}{ \begin{tabular}{@{}c@{}} Inertial-only Optimization \\ (10 KFs @ 4Hz) \end{tabular}} & \begin{tabular}{@{}c@{}} VI \\ BA  \end{tabular} &  \multicolumn{4}{c|}{ \begin{tabular}{@{}c@{}} VINS-Mono \\ Initialization \cite{qin2018vins} \end{tabular} } & \multicolumn{4}{c|}{ \begin{tabular}{@{}c@{}} Inertial-only Optimization \\ (10 KFs @ 10Hz) \end{tabular}} & \begin{tabular}{@{}c@{}} VI \\ BA \end{tabular} \\ 
\cline{2-5} \cline{6-9} \cline{11-14} \cline{15-18}

{} &  \multicolumn{2}{c|}{scale error (\%)} & & & \multicolumn{2}{c|}{scale error (\%)} & & & 4 Hz & \multicolumn{2}{c|}{scale error (\%)} & & & \multicolumn{2}{c|}{scale error (\%)} & & & 10 Hz \\
 \cline{2-3} \cline{6-7} \cline{11-12}  \cline{15-16} 

      \begin{tabular}{@{}c@{}} Seq. \\ Name \end{tabular} & MK  & \begin{tabular}{@{}c@{}} MK+ \\ BA1\&2 \end{tabular} & \begin{tabular}{@{}c@{}} $t_{Init}$ \\ (s) \end{tabular}    & \begin{tabular}{@{}c@{}} $t_{Tot}$ \\ (s) \end{tabular} &  \begin{tabular}{@{}c@{}}Inert. \\ Only \end{tabular}  & \begin{tabular}{@{}c@{}} Inert. \\ Only+BA \end{tabular} & \begin{tabular}{@{}c@{}} $t_{Init}$ \\ (s) \end{tabular} & \begin{tabular}{@{}c@{}} $t_{Tot}$ \\ (s) \end{tabular} & \begin{tabular}{@{}c@{}} scale \\ error (\%) \end{tabular} & \begin{tabular}{@{}c@{}}VI \\ Align. \end{tabular} & \begin{tabular}{@{}c@{}}VI Align. \\ + BA \end{tabular} & \begin{tabular}{@{}c@{}} $t_{Init}$ \\ (s) \end{tabular} & \begin{tabular}{@{}c@{}} $t_{Tot}$ \\ (s) \end{tabular} &  \begin{tabular}{@{}c@{}}Inert. \\ Only \end{tabular}  & \begin{tabular}{@{}c@{}} Inert.-Only \\ + BA \end{tabular} & \begin{tabular}{@{}c@{}} $t_{Init}$ \\ (s) \end{tabular} & \begin{tabular}{@{}c@{}} $t_{Tot}$ \\ (s) \end{tabular} & \begin{tabular}{@{}c@{}} scale \\ error (\%) \end{tabular} \\    \hline \hline 
        
V1\_01 & 21.11 & 5.39 & 2.23 & 3.18 & 10.41 & 4.99 & 2.16 & 2.78 & 15.04 & 50.2 & 27.67 & 1.15 & 1.54 & 20.34 & 7.69 & 1.26 & 1.89 & 26.04\\
V1\_02 & 31.37 & 6.11 & 0.97 & 3.79 & 11.27 & 4.48 & 2.15 & 2.88 & 21.59 & 69.65 & 36.36 & 1.46 & 2.46 & 29.98 & 9.83 & 1.27 & 1.82 & 19.95 \\
V1\_03 & 35.62 & 4.65 & 1.07 & 5.06 & 14.19 & 4.25 & 2.13 & 4.48 & 19.74 & 78.79 & 31.53 & 1.23 & 3.23 & 36.58 & 9.24 & 1.25 & 3.07 & 21.79\\

V2\_01 & 23.76 & 6.72 & 2.26 & 6.83 & 8.66 & 5.29 & 2.19 & 2.88 & 9.77 & 42.65 & 17.32 & 1.12 & 1.86 & 16.09 & 8.09 & 1.28 & 1.98 & 22.55\\
V2\_02 & 28.65 & 7.00 & 0.93 & 4.49 & 6.27 & 3.33 & 2.19 & 2.82 & 20.02 & 53.27 & 18.61 & 1.02 & 1.98 & 20.27 & 6.27 & 1.26 & 1.73 & 23.09\\
V2\_03 & 32.36 & 7.46 & 0.89 & 12.11 & 22.24 & 8.04 & 2.23 & 6.23 & 17.13 & 65.28 & 22.7 & 1.17 & 3.81 & 35.57 & 9.32 & 1.28 & 4.37 & 25.95\\
\hline 

MH\_01 & 29.23 & 7.65 & 2.98 & 12.87 & 6.21 & 4.48 & 2.14 & 5.03 & 11.78 & 20.98 & 16.86 & 2.78 & 3.08 & 18.12 & 6.2 & 1.26 & 3.97 & 18.90\\
MH\_02 & 21.62 & 5.71 & 2.93 & 10.57 & 7.07 & 4.31 & 2.14 & 3.77 & 18.02 & 19.89 & 13.15 & 1.64 & 2.03 & 19.76 & 6.82 & 1.27 & 2.61 & 23.08\\
MH\_03 & 28.75 & 6.38 & 2.06 & 15.51 & 9.88 & 4.77 & 2.16 & 3.32 & 5.27 & 39.3 & 16.09 & 1.30 & 1.75 & 31.32 & 9.89 & 1.25 & 2.29 & 10.34\\
MH\_04 & 28.65 & 5.23 & 2.41 & 36.57 & 16.3 & 7.86 & 2.14 & 3.59 & 6.77 & 57.19 & 20.28 & 1.11 & 2.11 & 43.78 & 13.78 & 1.26 & 2.82 & 11.15\\
MH\_05 & 25.28 & 3.51 & 2.73 & 38.00 & 16.05 & 6.37 & 2.17 & 3.48 & 6.84 & 55.6 & 21.96 & 1.34 & 1.72 & 41.64 & 12.88 & 1.26 & 2.4 & 12.13\\
\hline \hline
Mean Values	& 27.85 & 5.98 & 1.95 & 13.54 & 11.69 & 5.29 & 2.16 & 3.75 & 13.82 & 50.25 & 22.05 & 1.39 & 2.32 & 28.50 & 9.09 & 1.26 & 2.63 & 19.60\\
\hline

\end{tabular}
\end{table*}
\endgroup

Results are summarized in table \ref{tab:ResultsInit}. The first two blocks compare our method with our previous joint initialization method \cite{campos2019fast}, based on the work of Martinelli \cite{martinelli2014closed} and Kaiser et al. \cite{kaiser2017simultaneous}, improved by VI-BA and two rejection tests. The proposed initialization beats the joint initialization by a wide margin, both in accuracy and needed time, being able to initialize in less than 4 seconds with scale error of 5.29\%, using trajectories of 2.16 seconds on average. The method of \cite{campos2019fast} was able to obtain a scale error only slightly worse, but it was at the expense of a $t_{Tot}$ of 13 seconds, owing to the high rejection rate of the proposed tests. The baseline VI-BA initialization, using the same trajectories and initial guesses, obtains an average scale error of 13.82\%, which is even higher than the 11.69\% error obtained by just applying our inertial-only optimization, which is also much more efficient (5 ms per run, compared with 133 ms, as shown in table \ref{tab:ComputingTimeInit}).

The last blocks compare our method with the loosely-coupled initialization of VINS-Mono   \cite{qin2018vins}. To ease comparison, we have configured our system to run with similar-sized trajectories ($t_{Init}$) of around 1.25 seconds. With these shorter trajectories, our method beats the baseline VI-BA initialization doubling its accuracy and more than doubling the accuracy of VINS-Mono initialization, with a $t_{Tot}$ 0.31 seconds higher. This slightly higher $t_{Tot}$ is the result of the visual initialization used in our system, that one from ORB-SLAM, which in difficult sequences can struggle to success. We remark that reducing the scale error by using longer initialization trajectories, i.e. increasing $t_{Init}$, may not be easy for VINS-Mono. Since this system is not prepared to work as a pure visual odometry system, visual and inertial initializations have to be solved simultaneously for the same set of frames, and increasing time for inertial initialization would also increase the visual initialization time. This entails that points have to be tracked along more frames, which may be not feasible in case of camera rotation or fast motion.

For VINS-Mono, there is sharp contrast between the 22.05\% scale error found in our exhaustive initialization tests and the low RMS ATE error reported in \cite{qin2018vins} (in the range of 0.080-0.320\,m) when the whole trajectories are processed. This may be explained because when launched from the beginning, the initialization is performed while the drone is taking off, which entails big accelerations, making inertial parameters more observable, while in our experiment, initialization is performed along the whole sequence where other motions that give lower observability are present. Moreover, since VINS-Mono marginalizes old states, not fixing them as ORBSLAM-VI does, this initial error can be further reduced as the drone progresses.

\begin{figure}
  \centering
  \includegraphics[width=0.45\textwidth]{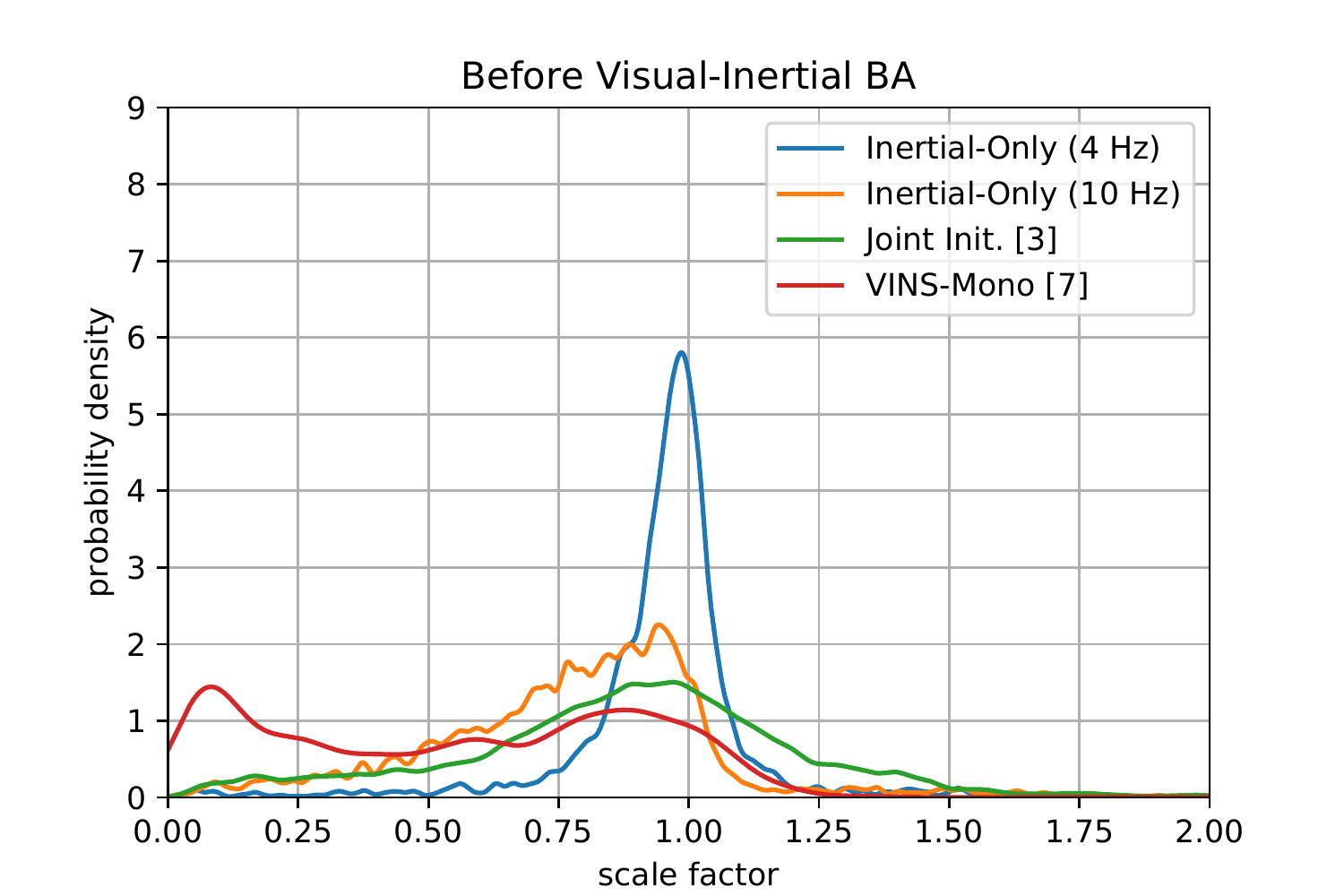}
  \includegraphics[width=0.45\textwidth]{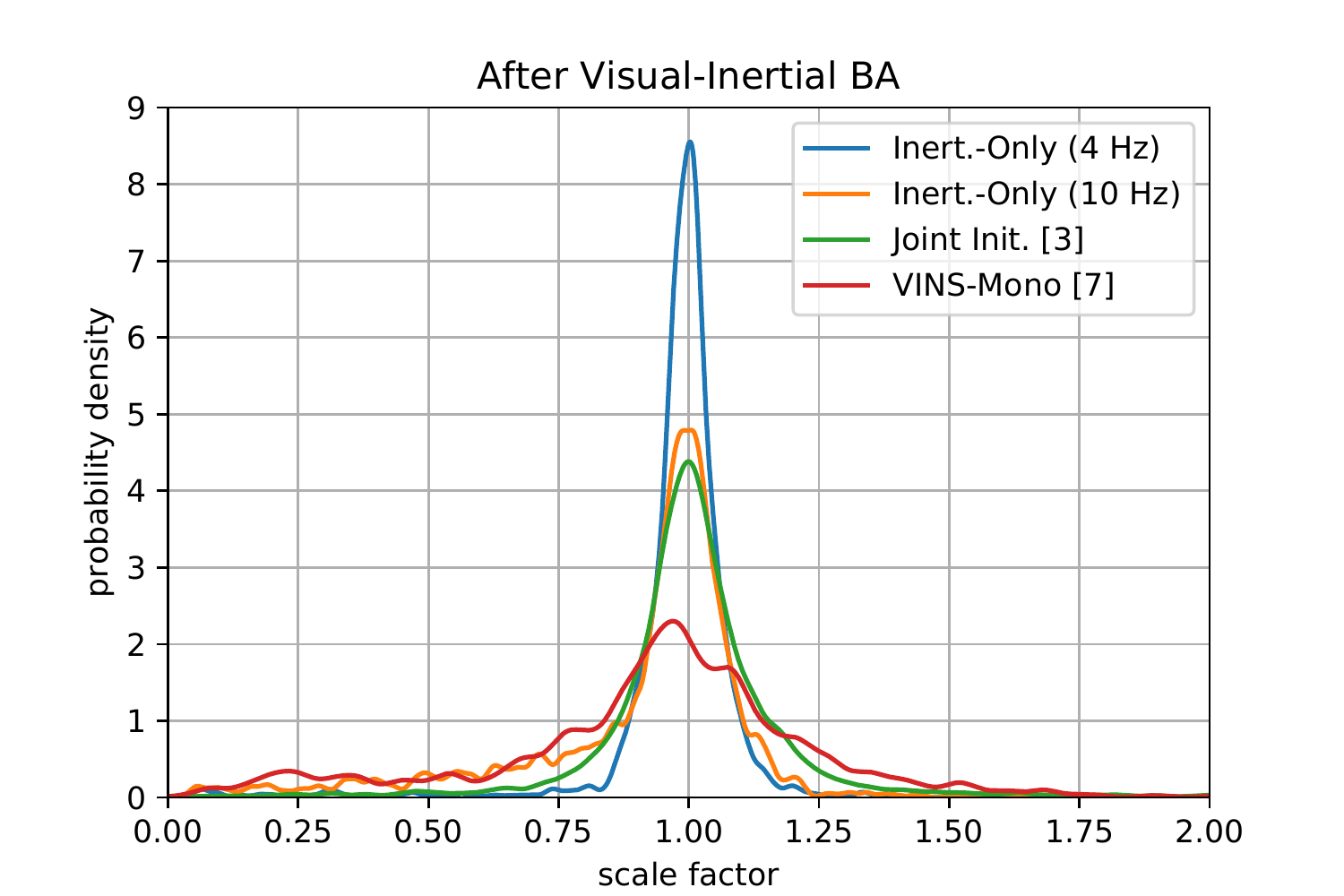}

  \caption{Experimental distribution of the scale factor (ratio between estimated and true scales) obtained by the different initialization methods along all sequences of the EuRoC dataset, before and after visual-inertial BA. A total of 2248 initializations have been launched.}
  \label{fig:scaleFactor}
\end{figure}

In figure \ref{fig:scaleFactor}, we plot the scale factor distribution for every studied method, along the whole EuRoC dataset. Results before visual-inertial BA show that all methods tend to underestimate the true scale. This bias is worse in VINS-Mono initialization, where there is a high number of initial solutions whose scale is close to zero. In contrast, the bias is much lower for inertial-only optimization at 4 Hz, that uses 2.15 s trajectories, whose mean is close to one. After visual-inertial BA, the bias almost disappears, having all methods a distribution with mean close to one, but with different variances, being VINS-Mono with 1s trajectories the worst and our inertial-only optimization with 2 s trajectories the best.

\begingroup
\begin{table}
\centering
\caption {\label{tab:ComputingTimeInit} Computing time of our method for the exhaustive initialization experiment in sequence V1\_02.}

\setlength\tabcolsep{3pt}
\begin{tabular}{|c|c|c|c|} \hline
Step & \text{mean (ms)} & \text{median (ms)} & \text{max (ms)} \\ \hline
Inertial-Only & 3$\times$5.24 & 3$\times$4.92 & 3$\times$6.39   \\
Map update & 11.18 & 11.25  & 13.98   \\
Visual-Inertial BA &  132.78 & 136.43 & 198.07 \\
 \hline \hline
 Total &  159.68 & 163.92 & 228.24 \\
\hline
\end{tabular}
\end{table}
\endgroup

Finally, considering the computing times in table \ref{tab:ComputingTimeInit}, inertial-only optimization is extremely efficient, taking around 5 ms per run, rotating and scaling points, frames and velocities takes 11 ms, and full VI-BA requires 132 ms. The inertial-only optimization time is much lower than the time required by the Martinelli-Kaiser closed-form solution, which is around 60 ms \cite{campos2019fast}. Compared with the baseline VI-BA which requires three runs with different scales for a total 400 ms, our complete  method only takes 160 ms, and doubles the scale accuracy.

Once verified that our inertial-only optimization performs better than previous initialization methods, we have made a second experiment which consists in launching ORB-SLAM Visual-Inertial \cite{mur2017visual} using our new initialization. As in \cite{campos2019fast}, we perform two visual-inertial bundle adjustment 5 and 10 seconds after initialization. We check three different sequences of EuRoC dataset, with different difficulty degrees, running 5 experiments for each one. We align both SLAM and GT trajectories, and Absolute Trajectory Error (ATE) is measured.

\begingroup
\begin{table}
\centering
\caption {\label{tab:SLAMnewInit} Results for ORBSLAM-VI with the proposed initialization (median values on five executions are shown) compared with results reported for original ORBSLAM-VI \cite{mur2017visual} and VINS-Mono in \cite{qin2019general}.}
\setlength\tabcolsep{3pt}
\begin{tabular}{cccccccc}
  & \multicolumn{2}{c}{\begin{tabular}{@{}c@{}} ORBSLAM-VI \\ + our \\ initialization \end{tabular}} & & \multicolumn{2}{c}{\begin{tabular}{@{}c@{}} ORBSLAM-VI \\ \cite{mur2017visual} \end{tabular}} & & \begin{tabular}{@{}c@{}} VINS-Mono \\ \cite{qin2019general} \end{tabular} \\
  \cline{2-3} \cline{5-6} \cline{8-8}
  \begin{tabular}{@{}c@{}} Seq. Name \end{tabular}  & \multicolumn{1}{c}{ \begin{tabular}{@{}c@{}} Scale \\ error (\%) \end{tabular}} &  \multicolumn{1}{c}{ \begin{tabular}{@{}c@{}} RMSE \\ ATE (m) \end{tabular}} & & \multicolumn{1}{c}{ \begin{tabular}{@{}c@{}} Scale \\ error (\%) \end{tabular}} & \multicolumn{1}{c}{ \begin{tabular}{@{}c@{}} RMSE \\ ATE (m) \end{tabular}} & & \begin{tabular}{@{}c@{}} RMSE \\ ATE (m) \end{tabular}  \\
  \hline
V1\_01 & 0.4 & 0.023 & & 0.9 & 0.027 & & 0.060\\
V1\_02 & 0.3 & 0.026 & & 0.8 & 0.024 & & 0.090\\
V1\_03 & 1.7 & 0.059 & & - & - & & 0.180\\
 \hline 
\end{tabular}
\end{table}
\endgroup

Results from table \ref{tab:SLAMnewInit} show that ORB-SLAM VI reaches in sequences V1\_01 and V1\_02 similar accuracy levels using the proposed initialization and the original initialization from \cite{mur2017visual}. In addition, sequence V1\_03, which previously could not be processed, because the original initialization failed, can now be successfully processed. This is because the new initialization takes just 2 seconds, being possible to immediately use the IMU, avoiding tracking loss during subsequent fast motions. Our results show that the combination of our initialization method with ORB-SLAM VI gives a very robust system that is significantly more accurate than VINS-Mono.

\section{Conclusions and Future Work}
The proposed initialization method has shown to be more accurate than the top-performing methods in the literature, with a very low computing time. This confirms that optimal estimation theory is able to make proper use of the probabilistic models of sensor noises, obtaining more accurate results than solving linear systems of equations or using non-weighted least squares. 

Full visual-inertial BA is a very non-linear problem, plagued with local minima, which hinders convergence. We have split it in a fully observable up-to-scale visual problem, followed by an inertial-only optimization phase that can be solved very efficiently, producing an initial solution for VI-BA that alleviates the local minima problem. 

As future work, we highlight that this inertial-only optimization could be used not only for initialization, but also to refine the scale and other inertial parameters once SLAM is initialized and running. This would have a much lower computational cost than performing a full visual-inertial bundle adjustment, where all visual and inertial parameters are involved. Moreover, this new initialization can be easily adapted to the stereo-inertial case. It would be enough to remove the scale from the inertial-only optimization.

\appendix

\section{Derivatives for inertial-only optimization}
Derivatives w.r.t. $\delta \mathbf{b}^g$, $\delta \mathbf{b}^a$, $\delta \mathbf{\bar{v}}_i$ and $\delta \mathbf{\bar{v}}_j$ are found or immediately derived from \cite{forster2015imu}. Derivatives for $\delta s$ are:
\begin{equation}
\frac{\partial \textbf{r}_{\Delta \text{R}_{ij}}}{\partial \delta s} = \mathbf{0}_{3 \times 1}
\end{equation}

\begin{equation}
\frac{\partial \textbf{r}_{\Delta \text{v}_{ij}}}{\partial \delta s} = \textbf{R}_i^\text{T} \left(\bar{\textbf{v}}_j - \bar{\textbf{v}}_i \right) s \exp(\delta s)
\end{equation}

\begin{equation}
\frac{\partial \textbf{r}_{\Delta \text{p}_{ij}}}{\partial \delta s} = \textbf{R}_i^\text{T} \left( \bar{\textbf{p}}_j - \bar{\textbf{p}}_i - \bar{\textbf{v}}_i \Delta t_{ij} \right) s \exp(\delta s)
\end{equation}

All these expressions are evaluated for $\delta s = 0$. Derivatives for $\delta \mathbf{g}_{dir}$ are:
\begin{equation}
\frac{\partial \textbf{r}_{\Delta \text{R}_{ij}}}{\partial \delta \mathbf{g}_{dir}}  = \mathbf{0}_{3 \times 2}
\end{equation}

\begin{equation}
\frac{\partial \textbf{r}_{\Delta \text{v}_{ij}}}{\partial \delta \mathbf{g}_{dir}} = -\textbf{R}_i^\text{T} \textbf{R}_{\text{wg}} \textbf{G} \Delta t_{ij}
\end{equation}

\begin{equation}
\frac{\partial \textbf{r}_{\Delta \text{p}_{ij}}}{\partial \delta \mathbf{g}_{dir}} = -\frac{1}{2} \textbf{R}_i^\text{T} \textbf{R}_{\text{wg}} \textbf{G} \Delta t_{ij}^2
\end{equation}

Where:
\begin{equation}
 \mathbf{G} =  \left(  \begin{array}{cc}
    0 & -G \\
    G & 0 \\
    0 & 0
\end{array} \right)
\end{equation}

\addtolength{\textheight}{-16.6cm}   

\bibliographystyle{IEEEtran}
\bibliography{VI_initialization,IEEEabrv}

\end{document}